\documentclass[11pt, oneside]{article}
\usepackage[utf8]{inputenc}
\usepackage[margin=1in]{geometry}
\usepackage{titlesec}
\usepackage{authblk}
\usepackage{adjustbox}
\usepackage[font=small,skip=5pt]{caption}
\usepackage{enumitem}
  \setlist[enumerate]{itemsep=0mm}
  \setlist[itemize]{itemsep=0mm}
\usepackage{amssymb, amsmath, amsthm}
  \DeclareMathOperator*{\argmax}{argmax}
  
\usepackage{mathtools}
\usepackage{algorithm}
\usepackage[noend]{algpseudocode}
\usepackage{listings}
\usepackage{bm}
\usepackage{todonotes}
\usepackage{subcaption}
\usepackage{graphicx}
  \graphicspath{ {images/} }
\usepackage{float}
\usepackage{url}
\usepackage{hyperref}
\hypersetup{
  citecolor=blue,
  linkcolor=blue,
  colorlinks=true,
  urlcolor=blue
}

\usepackage[backend=biber, style=numeric-comp, url=false, doi=false,
            isbn=false, eprint=false, natbib=true, maxbibnames=20,
            maxcitenames=2, sorting=none, uniquelist=false]{biblatex}

\AtEveryBibitem{\clearfield{month}}
\AtEveryBibitem{\clearfield{day}}
\AtEveryBibitem{\clearfield{version}}

\title{Deep reinforcement learning for irrigation scheduling using high-dimensional sensor feedback}

\author[1]{Yuji Saikai\thanks{ysaikai@unimelb.edu.au}}
\author[2]{Allan Peake}
\author[3]{Karine Chenu}
\affil[1]{School of Mathematics and Statistics, the University of Melbourne, Australia}
\affil[2]{CSIRO Agriculture (currently, Meat and Livestock Australia), Australia}
\affil[3]{The University of Queensland, Queensland Alliance for Agriculture and Food Innovation, Australia}

\date{}

\begin{document}


\maketitle
\begin{abstract}
\noindent
Deep reinforcement learning has considerable potential to improve irrigation scheduling in many cropping systems by applying adaptive amounts of water based on various measurements over time. The goal is to discover an intelligent decision rule that processes information available to growers and prescribes sensible irrigation amounts for the time steps considered. Due to the technical novelty, however, the research on the technique remains sparse and impractical. To accelerate the progress, the paper proposes a principled framework and actionable procedure that allow researchers to formulate their own optimisation problems and implement solution algorithms based on deep reinforcement learning. The effectiveness of the framework was demonstrated using a case study of irrigated wheat grown in a productive region of Australia where profits were maximised. Specifically, the decision rule takes nine state variable inputs: crop phenological stage, leaf area index, extractable soil water for each of the five top layers, cumulative rainfall and cumulative irrigation. It returns a probabilistic prescription over five candidate irrigation amounts (0, 10, 20, 30 and 40 mm) every day. The production system was simulated at Goondiwindi using the APSIM-Wheat crop model. After training in the learning environment using 1981--2010 weather data, the learned decision rule was tested individually for each year of 2011--2020. The results were compared against the benchmark profits obtained by a conventional rule common in the region. The discovered decision rule prescribed daily irrigation amounts that uniformly improved on the conventional rule for all the testing years, and the largest improvement reached 17\% in 2018. The framework is general and applicable to a wide range of cropping systems with realistic optimisation problems.

\vspace{10pt}
\end{abstract}

\vspace{10pt}
\section{Introduction}
Fresh water is becoming a scarce resource in many parts of the world, and its use in agriculture increasingly needs to be optimised. While there are a number of approaches to irrigation optimisation, irrigation scheduling using advanced sensor technologies has considerable potential to apply the right amount of water at the right time based on monitored plant, soil, and atmospheric conditions \citep{abioye_review_2020}. In operationalising precision irrigation, a significant challenge is to devise an intelligent decision rule that prescribes a sensible irrigation amount at the time of each decision making based on inputs from a variety of crop and environmental measurements \citep{jimenez_survey_2020}.

While precision irrigation, as a form of precision agriculture, holds the promise to increase resource use efficiency by exploiting advanced technologies, it is also faced with the challenge---the technologies are too complicated to fully exploit in practice \citep{lindblom_promoting_2017,saikai_machine_2020}. For example, drip irrigation is a prototypical practice of precision irrigation, enabling precise control of irrigation rate and timing. For determining rates and timings, \citet{abioye_review_2020} listed 18 basic parameters that can be readily monitored using available sensor devices: 10 crop parameters including leaf area index and sap flow, 4 soil parameters including soil moisture and salinity, and 4 weather parameters including temperature and rainfall. The question is, given the data stream generated by a variety of sensors, how to sensibly determine when and how much to activate the drip irrigation system. In other words, the task is to devise an intelligent decision rule that sequentially prescribes irrigation amounts based on high-dimensional sensor feedback in order that the irrigated amounts are collectively sensible to achieve overall production goals such as profit maximisation.

As if reflecting the difficulty of the task, most studies have been addressing irrigation optimisation problems using only low-dimensional sensor feedback and often focus on only irrigation timings without thorough consideration of irrigation amounts. For example, the vast majority of studies considers a single source of feedback at a time: soil moisture or soil water deficit \citep{jimenez_cyber-physical_2020}. In addition, irrigation rules often handle only timings and use irrigation amounts needed to replenish the soil to field capacity \citep{jimenez_cyber-physical_2020}. If soil moisture is the only feedback information and production goals are yield maximisation, then it is intuitive and probably reasonable to assume that replenishment of soil water is likely to reach the aim. However, the assumption here ignores the availability of other sources of feedback information. Moreover, real-world optimisation problems are rarely as simple as unconstrained yield maximisation. For instance, in regions with water restrictions, applying deficit irrigation to large cropping areas is often more profitable for the farm than focusing on fully irrigated crops for only small areas \citep{fereres_deficit_2007, peake_effect_2018}.

A standard method to find optimal decision rules in real-world systems is model predictive control (MPC), which has also been adopted in agricultural decision problems \citep{ding_model_2018} including irrigation scheduling \citep{abioye_model_2021}. Crucially, MPC requires mathematical models of how systems evolve over time. While, in many physical systems, models of dynamics can be derived based on Newtonian first principles, there are in general no such first principles for complex agricultural systems \citep{li_kriging-based_2021} due to the significant non-linearity in responses and states that characterise these systems \citep{altieri_agroecology_2018,gliessman_agroecology_1990}. Therefore, to apply MPC, dynamics needs to be estimated using data (i.e., system identification), which is feasible only in low-dimensional cases. Indeed, ``most of the existing works on system identification are based on the soil moisture equation without capturing the changing dynamics of soil, plant, and weather'' \citep[2]{abioye_model_2021}. This again ignores the availability of high-dimensional sensor feedback.

Reinforcement learning (RL) is a subfield of machine learning, which intends to discover intelligent decision rules without prior knowledge of systems dynamics \citep{sutton_reinforcement_2018}. As a form of machine learning, RL relies on data that encodes key information of decision-making experience in an environment of interest over time. Relevant data include both feedback from the decision environment (e.g., monitored crop, soil, and weather) and actions taken by following decision rules (e.g., irrigation amounts). In cases of high-dimensional feedback, intelligent decision rules are most likely complex functions that can capture intricate relations between feedback and appropriate actions. A standard approach to learning such complex functions from data is to take advantage of the representational power of deep learning \citep{lecun_deep_2015}. Hence, deep RL has emerged as a promising approach to discovering intelligent decision rules using high-dimensional feedback \citep{mnih_human-level_2015}.

Despite the tremendous potential for irrigation scheduling, the research on RL applications remains relatively sparse and mostly impractical. This is in contrast to the widespread adoption of supervised machine learning methods including deep learning \citep{kamilaris_deep_2018} and reflects the technical novelty of deep RL in the agricultural community. Until recently, there existed only a handful of RL studies in the literature \citep{bergez_comparison_2001,sun_reinforcement_2017}. While, over the past few years, some researchers started to apply deep RL to specific problems, their implementations involve several impractical components, making it difficult for other researchers to apply the methods to different problems. In RL, decision rules are discovered through trials and errors in learning environments. Thus, when environments are created by simulations, discovered rules are useful for real-world deployment only to the extent that simulated environments capture key dynamics of the real-world systems.

For this reason, some studies \citep{ding_drlic_2022,alibabaei_irrigation_2022} have little practical relevance because they use learning environments whose dynamics are empirical models directly estimated using historical observations. Importantly, to ``discover'' good decision rules, RL agents need to stumble across unseen states presented by environmental extrapolation, which is empirical models are known to be poor at. In contrast, \citet{chen_reinforcement_2021} constructs a problem-specific mechanistic model of systems dynamics by combining several component processes. Although preferable for extrapolation and possibly adequate for simple environments, in general, it is costly in time and other resources to manually construct high-fidelity models of complex agricultural systems, which may require thousands of component processes. Since distinct farms/crops necessitate construction of distinct environments, use of crop simulators is more scalable and practical as it allows researchers to create pertinent environments characterised by crop types, soil, weather, and management practices for their unique problems.

While \citet{yang_cropping_2020} use AquaCrop simulator \citep{steduto_aquacropfao_2009}, the constructed environment model has unrealistic features that undermine the practical relevance of the study. For example, they use a fixed weather pattern in every simulation, despite the fact that weather is the most important random aspect in crop production. Moreover, they use in-season yield estimates from the simulator as performance signals to facilitate the learning. In reality, such estimates are unreliable, especially at early stages. Practical RL methods avoid relying on unrealistic extra information and try to overcome the common challenge of sparse performance signals \citep{vecerik_leveraging_2017}. \citet{kelly_assessing_2022} also use AquaCrop and simulate a number of state variables that can be observed in practice. However, their learning procedure consists of ad hoc steps and configurations (e.g., optimising the decision interval to strictly one of 1, 3, 5 and 7 days, which may be too rigid to be practical). Consequently, the generality of its findings is questionable.

To accelerate the research on deep RL for irrigation scheduling, this paper provides a principled framework and actionable procedure that facilitate individual applications of deep RL using high-dimensional sensor feedback. The framework consists of a formal mathematical formulation of an optimisation problem (Section \ref{sec:problem}), a solution algorithm (Section \ref{sec:algorithm}), and a procedure for constructing learning environments and implementing the algorithm (Section \ref{sec:procedure}). In describing the procedure, key aspects of specifying both learning environments and learning algorithms are emphasised. To demonstrate effectiveness of the framework, a simulation study was conducted with the APSIM-Wheat crop model for irrigated wheat at Goondiwindi, Australia. A profit-maximising decision rule was learned in the simulated environment using 1981--2010 weather data, and it was tested using 2011--2020 weather data. The resulting profit for each of the testing years was compared against the benchmark profit obtained using an irrigation schedule optimised specifically for that particular year (Section \ref{sec:demo} and \ref{sec:results}). Finally, the discussion includes analysis of the case study, key assumptions, limitations, and future directions of the framework (Section \ref{sec:discussion}).

\section{Materials and methods}
\subsection{Irrigation optimisation problem}\label{sec:problem}
Suppose that irrigation management starts at time \(t=0\) (e.g., day of sowing) and ends at time \(t=T-1\) (e.g., day of reaching a certain growth stage). The unit of time \(t\) may be a day, several days, or a week. At each time step \(t\in\{0,1,\dots,T-1\}\), the environment provides a state of the system \((S_t)\). A decision rule uses the observed state as inputs/feedback and prescribes an action \((A_t)\) that specifies an irrigation amount for time step \(t\). After taking action \(A_t\), the environment also provides a ``reward'' \((R_{t+1})\), a net benefit of taking action \(A_t\) at state \(S_t\). The process continues until \(t=T-1\) and, once \(S_{T}\) is reaches, it moves into a special terminal state \((S^+)\) and the environment provides a terminal reward \((R^+)\). This sequence of state observations, actions, and rewards forms an episode of RL. The quantity that the decision-making agent tries to maximise is the sum of rewards:
\[
\sum_{t=0}^{T-1}R_{t+1} + R^+ = R_1 + R_2 + \dots + R_T + R^+.
\]
In irrigation optimisation contexts, rewards \(R_1, R_2,\dots,R_T\) may be costs of irrigation (i.e., negative benefits) and a terminal reward may be a crop revenue realised at the end of season.

Since a reward at \(t+1\) depends on a state and an action at \(t\), \(R_{t+1}\) is technically a function of \(S_t\) and \(A_t\), i.e., \(R_{t+1} = R(S_t,A_t)\). So, the quantity to maximise can be rewritten as follows:
\begin{equation}\label{obj}
\sum_{t=0}^{T-1} R(S_t,A_t) + R^+ = R(S_0,A_0) + R(S_1,A_1) + \dots + R(S_{T-1},A_{T-1}) + R(S^+,\cdot).
\end{equation}
Notice that no action is taken at \(S_{T}\) and \(S^+\), and terminal reward \(R(S^+,\cdot)\) is independent of actions.

Let \(\pi\) denote a stochastic decision rule, a mathematical function that maps a state \((s)\) to a probability of taking an action \((a)\), i.e., \(\pi(a,s) = \mathbb{P}(A=a|S=s)\) where \(A\) and \(S\) are random variables. In other words, decision rule \(\pi\) probabilistically prescribes irrigation amount \(a\) given state observation \(s\). Since different \(\pi\) prescribes a different action at each step and leads to a different sequence of rewards, the sum of rewards at the end of episode is dependent on \(\pi\). Note that, by assigning $1$ to a particular action, decision rules can be deterministic.

Using \(\pi\), the ultimate goal of learning can be formally expressed as solving the following maximisation problem:
\[
\max_\pi \mathbb{E}\left[\sum_{t=0}^{T-1} R(S_t,A_t) + R^+\right],
\]
where \(A_t\) is chosen according to \(\pi\) and \(\mathbb{E}\) is the expectation with respect to the randomness in both states and decision rule. Equivalently, the goal is to discover an optimal decision rule \(\pi^*\) such that
\begin{equation}\label{prob}
  \pi^* \in \argmax_\pi \mathbb{E}\left[\sum_{t=0}^{T-1} R(S_t,A_t) + R^+\right].
\end{equation}
Importantly, terminal reward \(R^+\) depends on \(A_{T-1}\) through \(S_{T}\) and \(S^+\), thereby depending on \(\pi\).

When solving complex optimisation problems using deep RL, it is computationally infeasible to find exact \(\pi^*\). To illustrate this, suppose five candidate actions to choose from at each of 150 time steps. The total number of possible action sequences is \(5^{150} (\approx 10^{105})\), which is too large for any existing computer to find an optimal sequence from. Consequently, a practical goal is to approximately solve the problem (\ref{prob}) by finding a sub-optimal yet good enough decision rule \(\pi^\dagger\) such that
\begin{equation}\label{subopt}
  \mathbb{E}\left[\sum_{t=0}^{T-1} R(S_t,A_t^\dagger) + R^+\right] \approx \mathbb{E}\left[\sum_{t=0}^{T-1} R(S_t,A_t^*) + R^+\right],
\end{equation}
where \(\pi^\dagger(a,s) = \mathbb{P}(A_t^\dagger=a|S_t=s)\) and \(\pi^*(a,s) = \mathbb{P}(A_t^*=a|S_t=s)\).

\subsection{Solution algorithm}\label{sec:algorithm}
As mentioned above, in practice, solving the maximisation problem (\ref{prob}) is equivalent to learning a good enough decision rule that prescribes an action for a given state at each time step. There are many RL algorithms that can discover equally good decision rules. While most of the existing studies adopt Q-learning, a popular class of algorithms, the framework presented in this paper assumes another class of algorithms called policy gradient, because it is conceptually simpler and helps to lower the technical barriers, whcih is important to accelerate the research on deep RL for irrigation optimisation. Recall that the goal of RL is to learn an optimal decision rule through trial-and-error experience. In Q-learning, however, what is improved throughout the learning is not a decision rule but instead a mathematical object called an action-value function, from which a decision rule is indirectly derived. In contrast, a policy gradient method explicitly maintains and directly improves a decision rule throughout the learning \citep{sutton_reinforcement_2018}. Among several variants of policy gradient, this paper adopts the classical algorithm REINFORCE \citep{williams_simple_1992}.

A description of the algorithm is provided below.

{
\centering
\begin{minipage}{.8\linewidth}
\begin{algorithm}[H]
\caption{REINFORCE}\label{alg:1}
\begin{algorithmic}[1]
  \State \textbf{require:} \(\pi, \theta_0, \alpha, N\)
  \For {\(n\in\{1,2,\dots,N\}\)}
    \State \(\theta \gets \theta_{n-1}\)
    \State Simulate an episode \(S_0,A_0,R_1,\dots,S_{T-1},A_{T-1},R_T\) by following \(\pi_\theta\)
    \State \(G \gets 0\)
    \For {\(t\in\{T-1,T-2,\dots,0\}\)}
      \State \(G \gets G + R_{t+1}\)
      \State \(\theta \gets \theta + \alpha G \nabla \log\pi_\theta(A_t,S_t)\)
    \EndFor
  \State \(\theta_n \gets \theta\)
  \EndFor
  \State \textbf{return} Best \(\theta\) among \(\{\theta_1, \theta_2, \dots, \theta_N\}\)
\end{algorithmic}
\end{algorithm}
\end{minipage}
\par
\vspace{1em}
}

\noindent \(\pi_\theta\) denotes a specific decision rule parameterised by \(\theta\). In other words, within a class of decision rules \(\pi\), a particular decision rule is specified by \(\theta\), which is gradually changed throughout the learning process, and the goal of learning is to find \(\theta^\dagger\) that specifies a good decision rule \(\pi^\dagger = \pi_{\theta^\dagger}\). \(\alpha\) is a small positive number that controls how much \(\theta\) changes after each episode throughout the learning process. \(N\) is the total number of training episodes. \(G\) is an intermediate dummy variable. Finally, \(\log\) and \(\nabla\) denote respectively the natural logarithm and the gradient with respect to \(\theta\).

Note that this section immediately follows the problem formulation (Section \ref{sec:problem}) to highlight solution algorithms as a key component of the framework. However, a choice of algorithms depends on the nature of a specific problem, so these two components need to be jointly considered when applying the framework in practice. For example, the more complex a problem, the more sophisticated the required algorithm is. Details are discussed in Section \ref{sec:discussion}.

\subsection{Learning procedure}\label{sec:procedure}
As stated in the introduction, RL is a technique for discovering a good decision rule for the environment considered through decision-making experience. While environments can be real-world settings, in most cropping systems, it is infeasible to carry out a sufficient number of field trials (i.e., episodes) in the real world. Hence, in the current framework, environments are created by simulations with APSIM \citep{holzworth_apsim_2018}. Learning takes place in these simulated environments by following RL algorithms that try to solve optimisation problems.

When a learned decision rule is intended to be deployed or tested in a real-world system, it is vital to ensure that the simulated environment is of high fidelity, capturing salient features of the corresponding real-world situation; otherwise, the learned decision rule would be used in too distinct conditions to perform adequately. As noted in many studies (e.g., \citet{berghuijs_calibrating_2021, collins_improving_2021}), to create high-fidelity APSIM simulations for particular real-world scenarios, the following factors should be carefully specified:
\begin{itemize}
  \item meteorological information,
  \item soil characteristics,
  \item cultivar characteristics, and
  \item management practices (except irrigation schedule, which is actively optimised).
\end{itemize}

Next, the optimisation problem (Section \ref{sec:problem}) needs to be specified with respect to each of the following:
\begin{itemize}
  \item State variables \((S_t)\). Recall that state variables are inputs from various sensors that provide useful information for sequential decisions. Thus, selection of state variables from hundreds of APSIM simulated variables is based on usefulness, similarity, and monitoring capabilities in the corresponding real-world situation. Similarity means close correspondence between information provided by real-world measurement and information provided by APSIM simulated variables. For example, the crop stage variable in APSIM correspond to a physiological development stage that can easily be measured by skilled operators in the real world. Monitoring capabilities are often limited by physical and economic constrains. For example, the crop stage is updated and accessible on a daily basis in APSIM, whereas skilled operators may only measure it at key stages.
  \item Candidate actions \((A_t)\). Candidate actions are possible irrigation amounts over which a stochastic decision rule determines a probability distribution at each time of decision making. For example, suppose irrigation of 0, 20, and 40mm are candidate actions for stage 14.5 and a LAI of 0.17 at \(t=150\) \((S_{150}=(14.5,0.17))\). Then, a decision rule \(\pi(a,s)\) may prescribe
  \begin{align*}
    \pi(0,(14.5,0.17)) &= 0.8\\
    \pi(20,(14.5,0.17)) &= 0.12\\
    \pi(40,(14.5,0.17)) &= 0.08
  \end{align*}
  implying that, given the state \(S_{150}=(14.5,0.17)\), \(A_{150}=0\), \(A_{150}=20\), and \(A_{150}=40\) are chosen with probability \(0.8\), \(0.12\), and \(0.08\) respectively. Similar to the above constraints, capabilities of irrigator may limit possible candidate actions. For example, even if APSIM can precisely simulate each of candidate amounts \(\{0, 5, 10, 15, \dots\}\) (mm), a real-world irrigator may lack such precision. In this case, the candidate set should be chosen coarser (e.g., \(\{0, 10, 20, \dots\}\)).
  \item Reward characteristics \((R_{t+1})\). As mentioned in Section \ref{sec:problem}, a reward \(R_{t+1}\) is a net benefit of taking action \(A_t\) under state \(S_t\) at time \(t\). The reward function \(R(S_t,A_t)\) should be specified so that the sum of rewards over the irrigation management window from \(t=0\) to \(t=T-1\) is the quantity that the decision-making agent hopes to maximise. Aside from the example of water costs and crop revenue illustrated above, reward may alternatively consist of some water use efficiency metrics.
  \item Unit of time \((t)\), start time \((t=0)\), and end time \((t=T-1)\) of irrigation management. Although a day is the default unit in many cases as APSIM adopts it, different units can be chosen based on other practical considerations including state monitoring and irrigator capabilities. The start and end times should also reflect practicality in the intended real-world situation to help the learning algorithm return a sensible decision rule.
\end{itemize}

Following the specification of the optimisation problem, details of the learning algorithm needs to be specified, which is indicated by ``\textbf{require}'' at the top of Algorithm \ref{alg:1}. \(\theta_0\) is an initial parameter vector of \(\theta\). Since \(\theta\) typically consists of millions of parameters in deep neural networks, it is customary to initialise \(\theta_0\) by drawing random numbers from the standard normal distribution. \(N\) is chosen depending on computational resources. Since it is the total number of training episodes, the larger \(N\), the better learning results are. \(\alpha\) is heuristically set (e.g., \(10^{-7}\)) in tandem with the neural network architecture described below. \(\pi\) denotes a class of decision rules or a neural network architecture, which determines a set of possible functions (i.e., decision rules) that can be represented by \(\theta\). In other words, different architectures have different numbers of parameters and different ways to combine them, implying different sets of possible functions to represent. Recall that machine learning is equivalent to exploring for a good \(\theta\). Thus, overly simple architectures excessively restrict the set of possible decision rules and limit its performance, whereas overly rich architectures have too many parameters to discover good ones. This trade-off, also known as model selection or hyper-parameter tuning, is one of the most challenging aspects in machine learning applications. An implication is that, similar to a choice of algorithms (Section \ref{sec:algorithm}), a choice of architectures depends on the nature of a specific problem.

Finally, with all the specifications, the algorithm is implemented by integrating the code of deep RL into the code of APSIM written in C\#. The integration can be seamlessly accomplished owing to the open-source APSIM next generation \citep{holzworth_apsim_2018} and TensorFlow.NET \citep{chen_tensorflownet_2018}. As a reference, the code used for the case study (Section \ref{sec:demo}) is available on the website (\url{https://github.com/ysaikai/RLIR}).

\subsection{Case study}\label{sec:demo}
Effectiveness of the framework is demonstrated for a scenario of spring wheat production in subtropical Australia. Due to the randomness involved in weather patterns and action selections by the decision rule, for reproducibility of the results, the random number generation in C\# program is fixed at the beginning of simulations using command \texttt{Random(0)}. Below is a summary of key specifications, followed by subsections that contain details.

\begin{table}[H]
\begin{center}
\begin{tabular}{ l l }
  Weather data & 1981--2010 for learning and 2011--2020 for testing\\ \hline
  State variables & Phenological stage, LAI, \(\{\text{ESW}_d\}_{d=1}^5\), CuIrrig, CuRain\\ \hline
  Actions & 0, 10, 20, 30, 40 (mm of irrigation)\\ \hline
  Total episodes & 20,000\\ \hline
  Rewards & Water costs and revenue
\end{tabular}
\end{center}

(LAI: leaf area index, ESW$_d$: extractable soil water at soil layer $d$, CuIrrig: cumulative amount since sowing, CuRain: cumulative rainfall since sowing)
\end{table}

\subsubsection{APSIM specifications}
Simulations were conducted for irrigated spring wheat production in Goondiwindi, Australia. The irrigation system was assumed to be a centre-pivot system, which is capable of being automated to apply precise amounts of water, and is used by some farmers in the region for water-saving features. Weather data for 1981--2020 was obtained using SILO Patch point data \citep{jeffrey_using_2001}, among which 1981--2010 were used for learning and 2011--2020 were used for testing. Apart from the random weather realisations, all the other specifications (e.g., soil characteristics and management practices) were common in every simulation. Regarding soil characteristics, APSoil \#906 (Thallon) was used with an initial soil water profile full at 20\% of the plant available water capacity (PAWC). Specifically, the soil texture was clay and PAWC was 198 mm, implied by the following drained upper limit (DUL), crop lower limit (CLL), and bulk density (BD):

\begin{table}[H]
\begin{center}
\begin{tabular}{ r r r r }
  Depth (cm) & DUL (mm/mm) & CLL (mm/mm) & BD (g/cc)\\ \hline
  0-15 & 0.405 & 0.234 & 1.299\\
  15-30 & 0.407 & 0.258 & 1.359\\
  30-60 & 0.410 & 0.257 & 1.351\\
  60-90 & 0.405 & 0.259 & 1.365\\
  90-120 & 0.391 & 0.254 & 1.403\\
  120-150 & 0.319 & 0.271 & 1.594\\
  150-180 & 0.287 & 0.271 & 1.677\\
\end{tabular}
\end{center}
\end{table}

\noindent Variety Sunbri was sown at 30 mm sowing depth and 100 seeds/m\(^2\) density, with the sowing window between 25 April and 1 June. If 1 June was reached, 20 mm of water would be automatically applied. Fertilisation took place at sowing with 350kg/ha of nitrogen to ensure non-limiting N supply \citep{peake_effect_2018} and focus solely on the impact of irrigation scheduling on the profit. Since each APSIM simulation returns one profit figure, the notion of ``episode'' in RL simply corresponds to ``simulation'' in APSIM. For reproducibility, the complete \texttt{.apsimx} file is available on the \href{https://github.com/ysaikai/RLIR}{website}.

\subsubsection{Learning specifications}
Nine state variables were considered to provide useful information for irrigation decisions while realistic for regular monitoring in real-world situations:
\begin{itemize}
  \item crop phenological stage (\textit{Stage}),
  \item leaf area index (\textit{LAI}),
  \item extractable soil water (\textit{ESW}) at five different layers: 0-15, 15-30, 30-60, 60-90, and 90-120 (cm),
  \item cumulative irrigation amount since sowing (\textit{CuIrrig}), and
  \item cumulative rainfall since sowing (\textit{CuRain}).
\end{itemize}
\textit{LAI} and \textit{Stage} provide information about the crop status. \textit{ESW}s measured at five different layers provide information about the soil water available to the crop. Finally, \textit{CuIrrig} and \textit{CuRain} provide information about the past decisions and rainfall events respectively. In terms of observability, while \textit{CuIrrig} and \textit{CuRain} are clearly simple to record, the other state variables are also considered observable in many real-world scenarios. In practice, \textit{LAI} can be estimated by, for instance, remote sensing \citep{zheng_retrieving_2009}. In addition, \textit{Stage} that represents observable physiological development processes can be measured in the field by moderately skilled operators. \textit{ESW}s are also measurable in practice using an appropriately calibrated moisture monitoring device.

Regarding actions, the candidate irrigation amounts were assumed to be 0, 10, 20, 30, and 40 mm, i.e., \(A_t \in \{0, 10, 20, 30, 40\}\) for all \(t\in\{0,1,\dots,T-1\}\). The learning started on the day of sowing and ended on the day when \textit{Stage} reaches 85 (``soft dough'') during mid grain filling. Consequently, the learning window was random because the sowing day was influenced by the preceding precipitation, which changed from one season to another, and the plant growth was influenced by not only the season but also the stochastic decision rule, which continued to evolve in the course of learning.

Following the formulation of optimisation problems (Section \ref{sec:problem}), the reward function \(R(S,A)\) was defined so that the sum \(R_1+R_2+ \dots +R_T+R^+\) would be equal to the (partial) profit---crop revenue minus water costs, which was also used in other studies \citep{yang_deep_2020}. Formally, first, the terminal state \(S^+\) was defined to be the situation in which yield was realised. Then, for \(t\in\{0,\dots,T-1\}\),
\[
R_{t+1} = R(S_t,A_t) = -c \times A_t
\]
and, for \(S^+\),
\[
R^+ = R(S^+,\cdot) = p \times \text{Yield}.
\]
where \(c\) was the unit water cost (\$/mm) assumed to be \(c=0.6\) or \$60 AUD per megalitre, a low-cost water scenario that would incur energy costs in capturing water from a nearby river and then pumping it through an overhead irrigation system. \(p\) was the output price (\$/kg of grains) assumed to be \(p=0.25\) or \$250 AUD per tonne, a typical ``on-farm'' price for wheat grain in the region after the cost of transport to the nearest grain depot was deducted.

As indicated in Line 10 of Algorithm \ref{alg:1}, the best \(\theta\) was returned as an outcome of learning after completing all \(N\) episodes. To determine which \(\theta\) among \(\{\theta_1, \theta_2, \dots, \theta_N\}\) was the one that defines the best decision rule \(\pi^\dagger\), the average profit over the past 100 episodes was adopted as the performance metrics. It was the moving average of order 100, and the performance of \(\theta_n\) for \(n\in\{100,101,\dots,N\}\) was the average profit over episodes \(n-99, n-98, \dots, n\) (and, for \(n<100\), the average profit over episodes \(1,2, \dots, n\)). The reason for the moving average as opposed to a single profit figure was the randomness involved in weather patterns and action prescriptions. The best \(\theta\) was the one that the corresponding decision rule \(\pi_\theta\) produced the highest moving average profit over the total \(N=20000\) episodes.

\begin{figure}[H]
    \centering
    \includegraphics[width=\linewidth]{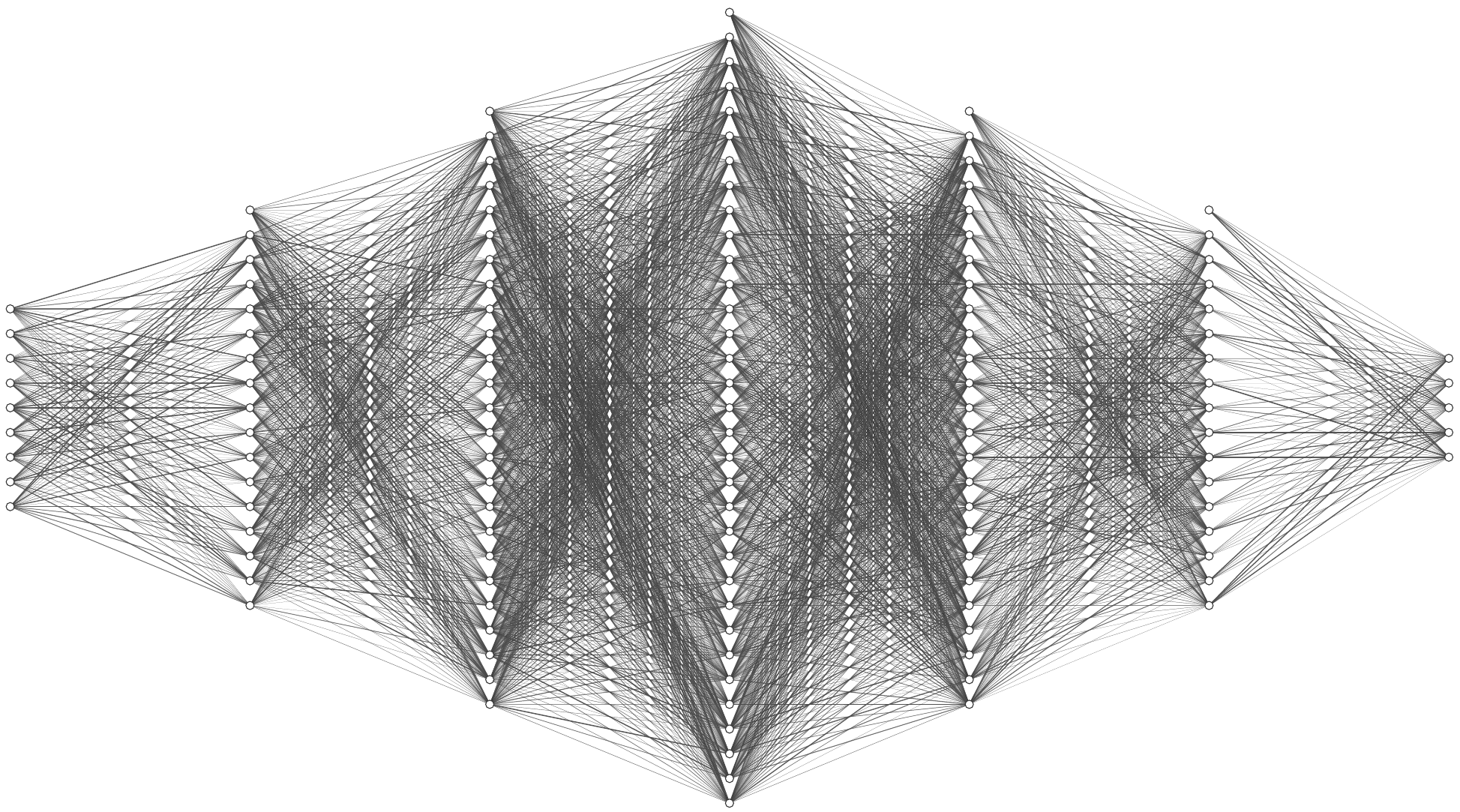}
    \caption{The neural network architecture used in the case study. The diagram indicates, for a given time \(t\), nine state variables (i.e., Stage, LAI, ESW of five soil layers, CuIrrig and CuRain) at the leftmost input layer and five candidate actions (i.e., irrigation of 0, 10, 20, 30 and 40mm) at the rightmost output layer. At every decision making, the neural network takes nine inputs from the sensors and returns five probabilities for action prescription. Due to the space restriction, only 4\% of the total number of nodes at each of five middle layers are drawn in the figure. A bias node (i.e., intercept term) is also drawn at the top of each of the middle layers.}
    \label{fig:architecture}
\end{figure}

The neural network used to model an irrigation decision rule was fully-connected and feed-forward, the most basic network architecture \citep{bishop_pattern_2006}. Specifically, it had five hidden layers in the middle, where 400 nodes in the first and fifth layers, 600 nodes in the second and fourth layers, and 800 nodes in the third layer as well as a single bias node at each layer (Figure \ref{fig:architecture}). Note that bias nodes in neural networks acted as intercept terms in linear functions. As indicated in the figure, the network took nine state variables as inputs, transformed them through the network, and output five numbers as action probabilities for five candidate irrigation amounts. With this architecture, the length of vector \(\theta\) that specified the decision rule \(\pi_\theta\) was equal to the number of edges in the network:

\begin{align*}
  1,448,005 ={} &(9\times400) + (400\times600+600) + (600\times800+800) +{} \\
              &(800\times600+600) + (600\times400+400) + (400\times5+5) .
\end{align*}

\noindent In other words, after each of 20,000 episodes, these 1,448,005 values were gradually changed in search of the best combination throughout the learning. As mentioned above, \(\alpha\) was heuristically searched for and set equal to \(10^{-7}\). The choice for activation functions also followed one of the most basic configurations---ReLU (rectified linear unit) \citep{nair_rectified_2010} at nodes in each hidden layer and ``softmax'' \citep{bishop_pattern_2006} at nodes in the output layer. Specifically, if the five unnormalised outputs of the neural network for five actions are \(f_0\), \(f_1\), \(f_2\), \(f_3\) and \(f_4\), the softmax activation for the output layer implies the following normalised numbers, i.e., irrigation probabilities, under the state vector equal to \(s\):
\[
\mathbb{P}(A_t=a_i|S_t=s) = \frac{e^{f_i}}{e^{f_0}+e^{f_1}+e^{f_2}+e^{f_3}+e^{f_4}},
\]
where \(a_i\in\{0,10,20,30,40\}\) for \(i\in\{0,1,2,3,4\}\) respectively, and \(e\) is the base of natural logarithm.

\subsubsection{Testing and benchmarking}
Once the training was completed, the learned decision rule was tested independently for each year of 2011--2020. Note that the decision rule learned in the environment based on the weather patterns of 1981--2010 performs well on average for these years but may be less optimal for 2011--2020. To assess its performance, two other sets of profits were also obtained. The first set was ``profit potential'', which is analogous to yield potential \citep{evans_yield_1999}, and used to approximate the maximum profit possible for each year. Specifically, to obtain the profit potential for a particular year, the same learning algorithm was run only in that year. Since the fixed year means a fixed weather pattern and the environment is no longer random, the search for the profit potentials is a simpler optimisation problem. Consequently, the profit potential approximates the maximum profit possible for each year.

The second set of profits was used as benchmark profits to assess the relative advantage of irrigation scheduling using high-dimensional sensor feedback. These are obtained by implementing a decision rule that reflects real-world recommendations in the case study region based on the soil moisture level, the most common low-dimensional sensor feedback. For the case study region, it is recommended that the soil moisture level should be maintained at 50--100 mm at the end of the season \citep{peake_better_2017}. Thus, to achieve this management objective, a replenishment rule was designed based on APSIM's \texttt{AutomaticIrrigation} management tool. Specifically, irrigation was automatically activated and the moisture level was restored to 95\% of PAWC, whenever the level dropped below 50\% of PAWC, which makes the trigger level of $99 \text{ mm} = 0.5\times198 \text{ mm}$.

\section{Results}\label{sec:results}
The decision rule was discovered by experimenting different irrigation amounts guided by the learning algorithm over 20,000 episodes, each of which was a random realisation of 30 years of wheat simulations (1981--2010) with the APSIM-Wheat crop model. The learned decision rule, mapping nine state variables (i.e. Stage, LAI, ESW of five soil layers, CuIrrig, and CuRain) to five action probabilities (i.e., probabilities to irrigate 0, 10, 20, 30 and 40 mm), is a complicated function characterised by 1,448,005 estimated neural network parameters (Figure \ref{fig:architecture}). Therefore, instead of trying to describe here mathematical properties of the decision rule, key features of the decision rule are highlighted by presenting summary statistics of the training and testing results as well as particular testing results for representative scenarios. All testing results are available on the \href{https://github.com/ysaikai/RLIR}{website}.

The learned decision ruled was tested 30 times for each year of 2011--2020. The reason for the replication is the following. For each day over the management window in a particular year, the decision rule takes nine state variables and prescribes five irrigation probabilities. Since an actual irrigation amount is randomly chosen according to these prescribed probabilities, there remains some randomness in resulting profits even for the same testing year with the fixed weather pattern. Thus, to reduce the effect of the randomness on the performance assessment, testing was conducted 30 times and the average profit was used as the final performance measurement for each testing year. Although 30 replicates may seem small, it turned out sufficient because the standard deviation of random profits was quite small (approximately \$10.8) relative to the average profit in every year. The testing result for each year is presented in Table \ref{table:results} together with the profit potential, the benchmark profit, and the relative performance against the benchmark.

\begin{table}[H]
\centering
\small
\caption{Testing results over 2011--2020. Test profits are the averages of 30 replicated testings of the learned decision rule for the corresponding years. Profit potentials are approximate maximum profits. Benchmark profits are ones obtained by the conventional replenishment rule. Performances are calculated as the ratios of the test profits over the benchmark profits.}
\label{table:results}
\begin{tabular}{r | *{4}{r}}
Year & Profit potential & Test profit & Benchmark & Performance\\
     & (\$/ha) & (\$/ha) & (\$/ha) & \\ \hline
2011 & 2,139 & \textbf{2,092} & 1,960 & 107\%\\
2012 & 2,197 & \textbf{2,164} & 1,973 & 110\%\\
2013 & 1,561 & \textbf{1,522} & 1,344 & 113\%\\
2014 & 1,972 & \textbf{1,932} & 1,753 & 110\%\\
2015 & 1,910 & \textbf{1,872} & 1,833 & 102\%\\
2016 & 1,799 & \textbf{1,778} & 1,656 & 107\%\\
2017 & 1,431 & \textbf{1,374} & 1,267 & 108\%\\
2018 & 1,556 & \textbf{1,513} & 1,289 & 117\%\\
2019 & 1,431 & \textbf{1,384} & 1,257 & 110\%\\
2020 & 1,711 & \textbf{1,673} & 1,614 & 104\%
\end{tabular}
\end{table}

\begin{figure}[H]
    \centering
    \includegraphics[width=\linewidth]{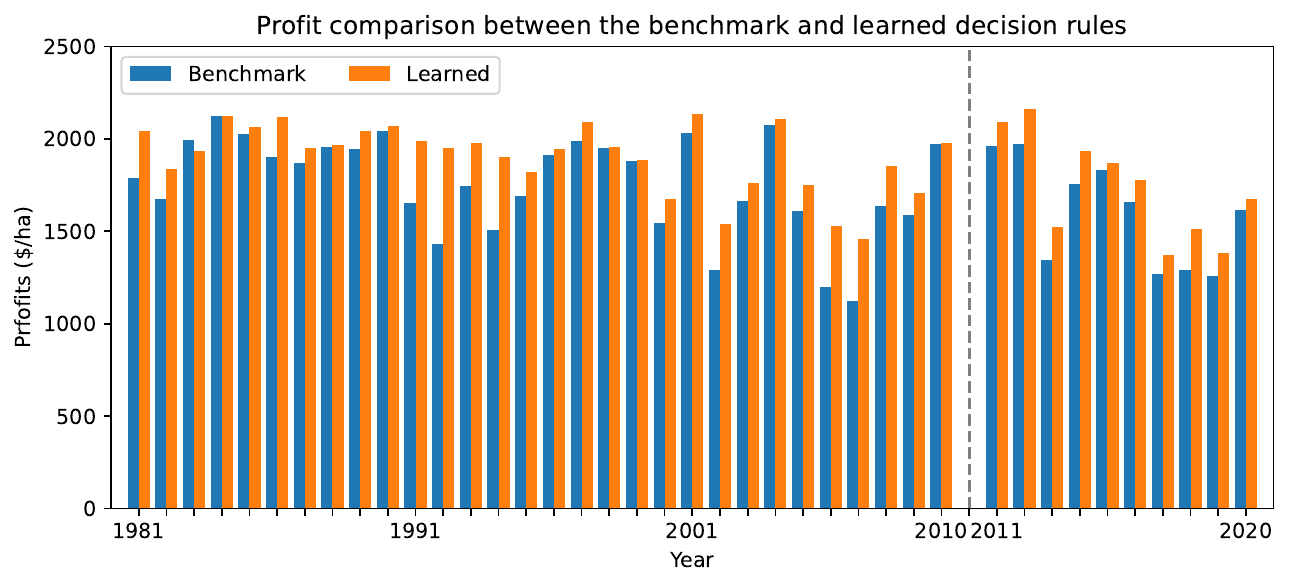}
    \caption{Profits resulting from the benchmark replenishment rule and the learned decision rule over the training (1981--2010) and the testing (2011--2020) years. Each profit of the learned decision rule is the average of 30 replicates.}
    \label{fig:profits}
\end{figure}

Overall, the learned decision rule managed to prescribe daily irrigation amounts that leads to at least 96\% of the corresponding profit potential in all testing years. The highest is 98.8\% in 2016, and the lowest is 96\% in 2017. More importantly, as seen by comparing the test and benchmark profits, the learned decision rule uniformly improves on the conventional replenishment rule. The largest improvement reaches 17\% in 2018. To provide a comprehensive comparison between the benchmark and learned decision rules, Figure \ref{fig:profits} plots all pairs of profits over the whole study period (1981--2020). Although the overall performance is satisfactory, as seen in 1983 and 1984 where the benchmark profits are higher, the utilisation of high-dimensional feedback is not perfect, reflecting the non-trivial nature of the learning task---discovering an intelligent rule that sequentially prescribes irrigation amounts based on daily feedback to maximise the profit at the end of the season.

\noindent For the sake of completeness, Table \ref{table:details} provides all the realised yield, cumulative rainfall, and cumulative irrigation amount prescribed by the learned decision rule.

\begin{table}[H]
\centering
\small
\caption{Yield (kg/ha), cumulative rainfall (mm), and cumulative irrigation (mm) under the learned decision rule over the whole study periods (1981--2020). Yield and cumulative irrigation in each year is the average over 30 replicates.}
\label{table:details}
\begin{tabular}{*{4}{r} | r *{4}{r}}
Year & Yield & Rainfall & Irrigation & & Year & Yield & Rainfall & Irrigation\\
     & (kg/ha) & (mm) & (mm) & & & (kg/ha) & (mm) & (mm)\\ \hline
1981 & 8,885 & 267 & 301 & & 2001 & 9,470 & 187 & 388\\
1982 & 8,293 & 130 & 393 & & 2002 & 7,295 & 130 & 474\\ 
1983 & 7,899 & 480 & 67 & & 2003 & 7,986 & 184 & 393\\ 
1984 & 8,877 & 286 & 161 & & 2004 & 9,242 & 193 & 339\\ 
1985 & 9,025 & 211 & 317 & & 2005 & 7,921 & 282 & 385\\ 
1986 & 9,244 & 213 & 325 & & 2006 & 7,332 & 107 & 504\\ 
1987 & 8,633 & 173 & 348 & & 2007 & 7,020 & 169 & 490\\ 
1988 & 8,211 & 298 & 141 & & 2008 & 8,364 & 158 & 396\\ 
1989 & 8,968 & 144 & 334 & & 2009 & 7,762 & 145 & 386\\ 
1990 & 8,919 & 184 & 265 & & 2010 & 8,032 & 395 & 47\\
1991 & 9,089 & 118 & 477 & & 2011 & 9,297 & 212 & 387\\ 
1992 & 9,102 & 101 & 545 & & 2012 & 9,667 & 126 & 421\\ 
1993 & 8,756 & 239 & 356 & & 2013 & 7,097 & 97 & 420\\ 
1994 & 8,809 & 80 & 499 & & 2014 & 8,796 & 103 & 446\\ 
1995 & 8,127 & 158 & 350 & & 2015 & 8,136 & 185 & 270\\ 
1996 & 8,328 & 257 & 229 & & 2016 & 7,646 & 296 & 223\\ 
1997 & 9,212 & 227 & 357 & & 2017 & 6,720 & 128 & 510\\ 
1998 & 7,908 & 537 & 32 & & 2018 & 7,164 & 156 & 463\\ 
1999 & 8,147 & 188 & 255 & & 2019 & 6,743 & 39 & 502\\ 
2000 & 7,834 & 78 & 472 & & 2020 & 7,493 & 137 & 334
\end{tabular}
\end{table}

\begin{figure}[H]
    \centering
    \begin{subfigure}{\linewidth}
        \includegraphics[width=\linewidth]{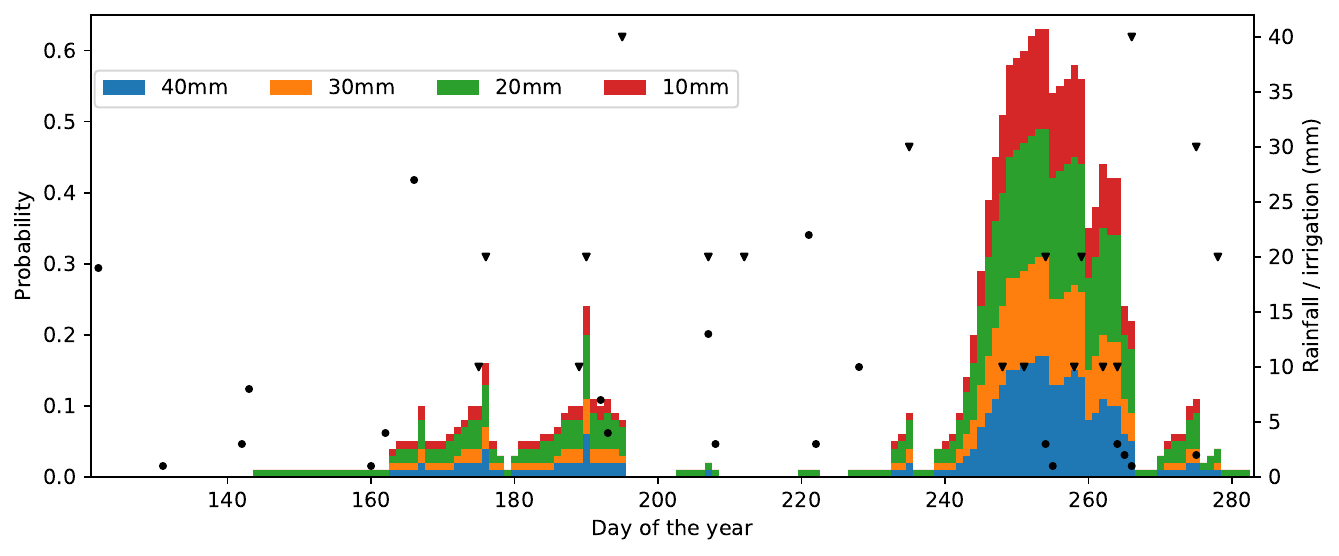}
        \caption{}
        \label{fig:probabilities}
    \end{subfigure}
    
    \begin{subfigure}{\linewidth}
        \includegraphics[width=\linewidth]{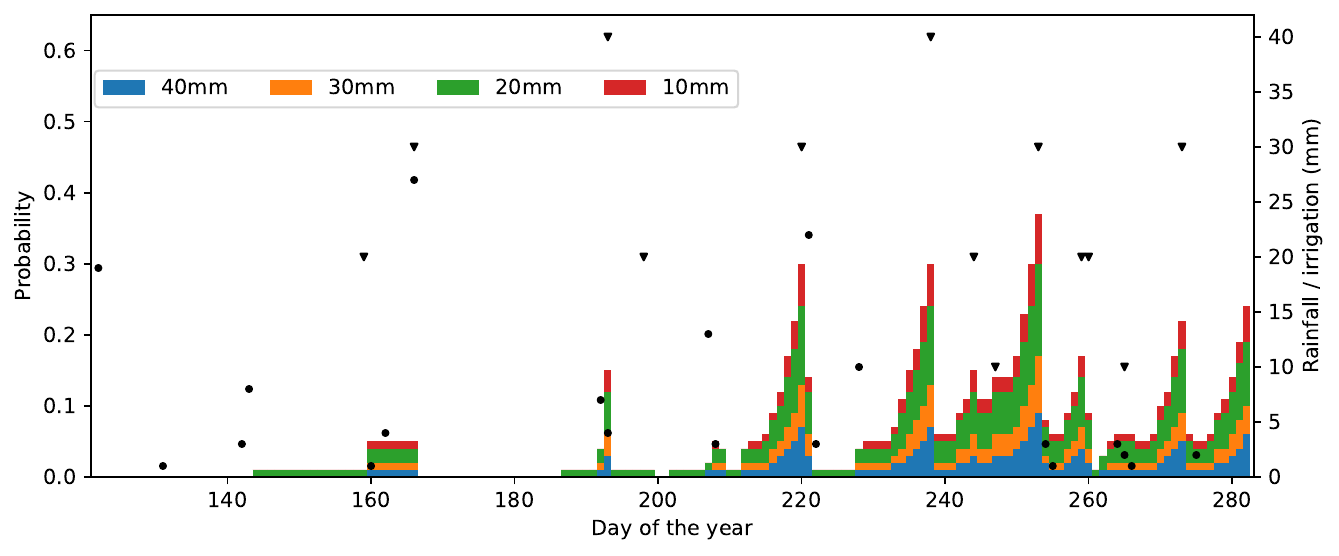}
        \caption{}
        \label{fig:probabilities2}
    \end{subfigure}
    
    \caption{Prescribed daily irrigation probabilities from sowing (Day 122) to stage 85 (Day 285) in two of 30 replicates for Year 2020. The first replicate is presented in (a) and the second one in (b). Dots (\(\bullet\)) and triangles (\(\blacktriangledown)\) represent daily rainfall and realised irrigation amounts respectively.}
\end{figure}

To gain some insight into how the learned decision rule makes daily prescriptions, Figure \ref{fig:probabilities} \& \ref{fig:probabilities2} illustrates a sequence of the prescribed action probabilities in two of 30 replicates for Year 2020. At each day, the probabilities for non-zero irrigation amounts are vertically stacked and colour-coded, and the realised irrigation that was applied is presented by a triangle. Since rainfall has a direct impact on irrigation decisions, daily rainfall is also included in the figures. Overall, in the first replicate (Figure \ref{fig:probabilities}), the positive prescriptions are significantly concentrated between Day 240 and 266. The resulting prescriptions were substantially different in the second replicate (Figure \ref{fig:probabilities2}) where the positive prescriptions are less concentrated and more evenly spread. However, despite the stark contrast in prescriptions, the resulting yields and total irrigation amounts are surprisingly similar (7,516 kg/ha v. 7,517 kg/ha using 350 mm and 320 mm of water, respectively), indicating the consistent performance of stochastic decision rule on average.

\noindent Again, it is important to keep in mind that an action prescription (i.e., probabilities for irrigation amounts) at any point in time strongly reflected and reacted to the underlying dynamics, which consisted of both the external states (e.g., daily climatic demand and soil status) and previous irrigations. As illustrated in Figure \ref{fig:probabilities} \& \ref{fig:probabilities2}, once a different irrigation amount was applied, the dynamics changed and led to distinct subsequent irrigation probabilities and realisations of states (e.g., soil water) despite the fact that daily climatic variables were fixed (i.e., the same climatic data used for replicates in the same year).

To illustrate the intricate processing of high-dimensional feedback, Table \ref{table:result2020} presents a sequence of state variables and prescriptions made by the decision rule during the period of intensive irrigation in the replicate corresponding to Figure \ref{fig:probabilities}. For instance, while ESW1 and ESW2 are quite similar between Day 240 and 258, the prescribed probabilities for no irrigation \(p(0)\) are significantly different, indicating soil moisture alone does not provide enough information. Another behaviour to note is that, aligned with the common intuition, \(p(0)\) tends to rise immediately after any positive amount of irrigation, while all the probabilities dynamically change throughout the season adapting to the dynamics of state variables. However, this is merely a tendency and not a fixed relation: after 10 mm of irrigation, \(p(0)\) falls by 7 percentage points on Day 249, remains unchanged on Day 252, and rises by 18 percentage points on Day 265. The complete data for all the replicates is available on the \href{https://github.com/ysaikai/RLIR}{website}.

\begin{table}[H]
\centering
\caption{Illustration of the complexity of the learned decision rule during the period of intensive irrigation in the replicate corresponding to Figure \ref{fig:probabilities}. Each \(p(A)\) indicates the probability for applying \(A\) amount of water (0, 10, 20, 30 or 40 mm) prescribed based on values of nine state variables (only six shown for the space restriction). ESW1 and ESW2 are similar on Day 240 and 258, but the probabilities for no irrigation \(p(0)\) are significantly different. After 10 mm of irrigation, \(p(0)\) falls by 7 percentage points on Day 249, remains unchanged on Day 252, and rises by 18 percentage points on Day 265.}
\label{table:result2020}
\small
\begin{tabular}{c|*{6}{c}|*{5}{c}}
 & \multicolumn{6}{|c|}{State variables} & \multicolumn{5}{|c}{Action probabilities}\\
Day & Stage &  LAI & ESW1 & ESW2 & CuIrrig & CuRain & \(p(0)\) & \(p(10)\) & \(p(20)\) & \(p(30)\) & \(p(40)\)\\
 &  &  & (mm) & (mm) & (mm) & (mm) &  &  &  & \\\hline
240 & 47.2 & 5.46 & \colorbox{black!10}{24} & \colorbox{black!10}{23} & 170 & 125 & \colorbox{black!10}{0.95} & 0.01 & 0.02 & 0.01 & 0.01\\
241 & 48.0 & 5.41 & 23 & 22 & 170 & 125 & 0.94 & 0.01 & 0.03 & 0.01 & 0.01\\
242 & 48.8 & 5.37 & 21 & 21 & 170 & 125 & 0.91 & 0.01 & 0.04 & 0.02 & 0.02\\
243 & 49.7 & 5.33 & 20 & 20 & 170 & 125 & 0.86 & 0.02 & 0.06 & 0.03 & 0.03\\
244 & 50.7 & 5.28 & 19 & 20 & 170 & 125 & 0.80 & 0.04 & 0.08 & 0.04 & 0.04\\
245 & 51.8 & 5.20 & 17 & 19 & 170 & 125 & 0.72 & 0.05 & 0.11 & 0.06 & 0.07\\
246 & 52.7 & 5.13 & 16 & 18 & 170 & 125 & 0.62 & 0.08 & 0.14 & 0.08 & 0.09\\
247 & 53.8 & 5.07 & 14 & 17 & 170 & 125 & 0.54 & 0.09 & 0.15 & 0.10 & 0.11\\
248 & 65.0 & 5.03 & 13 & 16 & 170 & 125 & \colorbox{black!10}{0.50} & 0.11 & 0.16 & 0.11 & 0.13\\
249 & 66.0 & 4.95 & 22 & 16 & 180 & 125 & \colorbox{black!10}{0.43} & 0.13 & 0.17 & 0.13 & 0.15\\
250 & 67.0 & 4.87 & 20 & 15 & 180 & 125 & 0.41 & 0.13 & 0.18 & 0.13 & 0.15\\
251 & 67.9 & 4.79 & 18 & 14 & 180 & 125 & \colorbox{black!10}{0.39} & 0.13 & 0.18 & 0.13 & 0.16\\
252 & 68.8 & 4.71 & 26 & 15 & 190 & 125 & \colorbox{black!10}{0.39} & 0.14 & 0.18 & 0.14 & 0.16\\
253 & 69.7 & 4.59 & 24 & 14 & 190 & 125 & 0.38 & 0.14 & 0.18 & 0.14 & 0.17\\
254 & 70.6 & 4.48 & 22 & 14 & 190 & 125 & 0.37 & 0.14 & 0.18 & 0.14 & 0.17\\
255 & 71.3 & 4.36 & 33 & 24 & 210 & 128 & 0.46 & 0.12 & 0.17 & 0.12 & 0.13\\
256 & 71.8 & 4.26 & 29 & 24 & 210 & 129 & 0.46 & 0.12 & 0.18 & 0.12 & 0.13\\
257 & 72.3 & 4.17 & 26 & 23 & 210 & 129 & 0.44 & 0.12 & 0.18 & 0.12 & 0.14\\
258 & 72.8 & 4.09 & \colorbox{black!10}{24} & \colorbox{black!10}{22} & 210 & 129 & \colorbox{black!10}{0.41} & 0.13 & 0.18 & 0.12 & 0.15\\
259 & 73.3 & 4.01 & 29 & 23 & 220 & 129 & 0.43 & 0.12 & 0.18 & 0.12 & 0.14\\
260 & 73.8 & 3.93 & 31 & 26 & 240 & 129 & 0.65 & 0.07 & 0.13 & 0.07 & 0.08\\
261 & 74.4 & 3.84 & 28 & 25 & 240 & 129 & 0.62 & 0.07 & 0.14 & 0.08 & 0.09\\
262 & 74.9 & 3.72 & 25 & 24 & 240 & 129 & 0.57 & 0.09 & 0.15 & 0.09 & 0.11\\
263 & 75.5 & 3.60 & 31 & 24 & 250 & 129 & 0.58 & 0.08 & 0.15 & 0.09 & 0.10\\
264 & 76.1 & 3.46 & 27 & 24 & 250 & 129 & \colorbox{black!10}{0.58} & 0.08 & 0.15 & 0.09 & 0.10\\
265 & 76.7 & 3.33 & 32 & 27 & 260 & 132 & \colorbox{black!10}{0.76} & 0.04 & 0.09 & 0.05 & 0.06\\
266 & 77.3 & 3.20 & 30 & 26 & 260 & 134 & 0.78 & 0.04 & 0.09 & 0.04 & 0.05\\
267 & 77.9 & 3.09 & 32 & 27 & 300 & 135 & 0.99 & 0.00 & 0.01 & 0.00 & 0.00
\end{tabular}
\end{table}

\section{Discussion}\label{sec:discussion}
\subsection{The case study}
The results from the case study suggest an alternative approach to irrigation scheduling in the region. For an irrigated wheat crop at Goondiwindi, growers typically make irrigation decisions using simple rules based on a phenological stage or soil moisture levels, which are monitored using capacitance probes installed 25 to 30 days after plant emergence. Even if more information is available, it will be virtually impossible for growers to process all the information at once and make sensible decisions every time. The case study demonstrated that, by algorithmically learning decision rules in the form of mathematical functions, it would be possible to make use of all the relevant information.

Comparing the simulated yields and irrigated amounts with the corresponding numbers in the existing studies, the learning environment constructed for the case study seems to be a reasonable one. Specifically, in this region, the maximum yield is estimated to range between 6.8 and 8.7 t/ha, utilising up to 550 mm of water including initial soil water, rainfall and irrigation \citep{sykes_irrigation_2012, peake_effect_2018}. The simulated yields varied between 7.0 and 9.5 t/ha for the learning years (1981--2010) and between 6.7 and 9.7 t/ha for the testing years (2011--2020). Over the whole studied period (1981--2020), the initial soil water was set to 40 mm, the average rainfall was 195 mm ranging between 39 and 537 mm, and the average amount of irrigated water was 350 mm ranging between 32 and 545 mm.

The following are additional comments on the specifications and results of the case study. The set of candidate actions was chosen to be \(\{0, 10, 20, 30, 40\}\) mm of irrigation because 40 mm per day was a reasonable maximum amount for the assumed use of overhead irrigators. However, greater irrigation amounts can be applied as done by some growers with other irrigation systems in the region. In terms of computational practicality, the total number of training episodes was set equal to 20,000 based on the available computational capacity. While it turned out sufficient for this particular scenario, different values may be adapted to other problems. Finally, for reproducibility of the case study, the random number generation in C\# program was fixed at the beginning of learning by choosing number 0 in the command \texttt{Random(0)}. Use of other numbers could lead to distinct learning results, but learned decision rules would likely perform equally well.


\subsection{Assumptions and limitations}
For successful applications of the framework, it is crucial to construct learning environments of high-fidelity; that is, creating environments that capture salient features of the corresponding real-world situations into which the learned decision rules are intended to be deployed. In the case study, the testing/deployment environment was the same as the learning environment. As a result, it was relatively easy to achieve high performance of the learned decision rule (i.e., achieving more than 96\% of the profit potential). In contrast, when deploying decision rules discovered by simulation into real-world systems, there are inevitable gaps between simulated and real-world environments, which will likely result in lower performance than those in the case study. Based on the existing knowledge about intended systems (e.g., crops, soil properties, and management practices), researchers need to narrow the gaps to claim the practical usefulness of discovered decision rules. The task is considered reasonable because of the existence of numerous studies that verify the fidelity of APSIM in a wide range of production systems especially for wheat in Australia \citep{lilley_seasonal_2007, chenu_environment_2011, christopher_stay-green_2016}.

Another important yet tacit assumption is that the randomness in testing/deployment environments is reasonably represented in learning environments. Recall the problem formulation (\ref{prob}), where the objective function to maximise is the expectation with respect to action selection and state distribution. When the learned decision rules are deployed without any modification, the randomness in the former is identical between learning and deployment environments. However, the randomness in the latter may differ. In the case study, the weather pattern at each training episode was randomly chosen from 30 possible patterns with probability 1/30. An implicit assumption was that the weather pattern in each of 10 testing years (2011--2020) was a realisation of the weather distribution collectively created by 30 patterns (1981--2010). Clearly, none of 10 patterns in the testing was exactly the same as one of 30 patterns in the learning. But, loosely speaking, each pattern in the testing was ``statistically similar'' to the created distribution. In practice, climate change may invalidate the assumption, necessitating separate creation of weather patterns for learning environments (e.g., fit time-series models to weather data and sample from the models). In addition, if other randomness is introduced to learning environments, researchers must ensure that it is also present in deployment environments.

Lastly, there is no consideration of spacial variations in the current framework; i.e., a single production environment is assumed for both learning and deployment. This stems from the current limitation of crop simulators. To obtain a decision rule capable of handling a spatial variation, researchers may create multiple environments according to the spatial variation, learn multiple decision rules through independent learning processes, and select an appropriate one based on spatial characteristics in the deployment environment each time of decision making.

\subsection{Future directions}
The optimisation problem in the case study was formulated as a simple one---the profit maximisation with the unlimited water availability. The simple formulation is deliberate in order to highlight the key implementation aspects of the deep RL framework, which is significantly more sophisticated than standard optimisation techniques. Since the proposed framework is general and flexible, it can be applied to a variety of real-world problems, which are often more complex. For example, a grower may manage multiple crops, instead of a single crop, and want to maximise the farm profit by optimising the farm-level water use on the whole. In this case, multiple learning environments can be constructed and run simultaneously. In another scenario, the total amount of water available to a grower may be constrained in each year or even across multiple years. In this case, the remaining water budget at any time step can be included as another state variable. Moreover, the framework can be used to optimise other management practices such as fertilisation together with irrigation. In this scenario, successfully learned decision rules can take into account the interaction between different management practices (e.g., nitrogen application and irrigation). In most scenarios, difference will be in specification of rewards and state variables.

Regarding learning algorithms (Section \ref{sec:algorithm}), the framework assumes the classical REINFORCE due to its conceptual simplicity, which helps to lower the technical barrier and allows applied researchers to focus on their applications. REINFORCE worked and discovered the decision rule that uniformly improved on the conventional replenishment rule for all the testing years in the case study (Table \ref{table:results}) which was characterised by nine state variables, five candidate actions, and the deterministic APSIM simulations (aside from the random weather pattern). If tackling more complex problems (e.g., more state variables, wider candidate actions, and greater randomness in learning environments), it will likely be necessary to use more advanced algorithms such as actor-critic methods \citep{sutton_reinforcement_2018} or even state-of-the-art algorithms. In addition, if deterministic decision rules as opposed to stochastic ones are preferred, it requires more advanced algorithms such as Deterministic policy gradient algorithms \citep{silver_deterministic_2014}. Nevertheless, main difference will be in implementation of algorithms, while the rest of the framework remains the same.

Use of weather forecast as a state variable is possible and simple to include in the framework, provided that some numerical values are available (e.g., expected amounts of rainfall over the following days or weeks). A key factor in deciding whether to include it is the quality of forecast. For example, when a significant amount of rainfall is reliably predicted in the coming days, a good decision rule will prescribe a high probability for no irrigation. By including reliable rain forecast in the framework, such decision rules can be automatically learned. However, if the forecast is unreliable, it may mislead the learning process and result in a poor decision rule.

Computational costs of learning depend on specifications of simulated environments, problem formulations, and learning algorithms. For example, if an optimisation problem is quite complex, requiring many state variables and an advanced learning algorithm, it will likely necessitate a large number of training episodes and long running time. As a reference, the learning over 20,000 episodes in the case study can be completed within 24 hours on a modern desktop computer.

Finally, to make the proposed framework truly practical and accelerate the research on RL in agriculture, it is vitally important to develop user-friendly software that dramatically reduces manual coding and facilitates implementation of the framework. In the current state of the software development, specific implementation requires some coding skills in C\# programming language for use of TensorFlow.NET and its integration into APSIM. Since the industry is rapidly increasing use of digital technologies in part with the help of high-tech developing and advising companies, collaboration across academia, public sectors, and the industry is possible. All the code used for the case study is available on the \href{https://github.com/ysaikai/RLIR}{website} and serves as a template that can be adapted to diverse practical problems.

\section{Conclusion}
This paper proposes a deep reinforcement learning (RL) framework for irrigation optimisation, which comprises a formal mathematical formulation of optimisation problems, a solution algorithm, and a procedure for constructing learning environments and implementing the algorithm. The effectiveness of the framework was demonstrated in the case study where the profit in wheat production was maximised by learning a near optimal irrigation schedule using deep RL. Specifically, the learning environment (1981--2010) was built using the APSIM-Wheat crop model. The learned decision rule was examined in the testing environment (2011--2020) and uniformly improved on the conventional replenishment rule. The proposed framework is flexible and can be used to address many complex problems (e.g., maximising water use efficiency or yield with a water budget). Since there remain technical barriers for some users (i.e., mathematical formulation and implementation using C\#), it is crucial to develop user-friendly software to facilitate applications of the framework. It is the authors' hope that this paper will spark wide collaboration across academia, public sectors and the industry to advance software development, help many practitioners solve their management optimisation problems, and collectively move towards economically and environmentally sustainable agriculture.


\printbibliography


\end{document}